\documentclass[11pt]{article}

\usepackage{amsmath, amsthm, amssymb, amscd, paralist}
\usepackage{graphicx, color}
\usepackage{bm}
\usepackage[title]{appendix}
\usepackage{hyperref}
\usepackage{wrapfig}
\usepackage{mdwlist,setspace,floatflt,rotating,url}
\usepackage{framed}

\newcommand{\wh}{\widehat}

\newcommand{\ol}{\overline}

\numberwithin{equation}{section}
\newtheorem{thm}{Theorem}

\newcommand{\E}{\mathbb{E}}

\newcommand{\mN}{\mathcal{N}}

\newcommand{\mT}{\mathcal{T}}
\newcommand{\mU}{\mathcal{U}}

\newcommand{\bu}{\mathbf{u}}
\newcommand{\bU}{\mathbf{U}}
\newcommand{\bv}{\mathbf{v}}

\newcommand{\bw}{\mathbf{w}}

\newcommand{\bX}{\mathbf{X}}

\newcommand{\bY}{\mathbf{Y}}

\newcommand{\bxi}{\bm{\xi}}

\newcommand{\btheta}{\bm{\theta}}

\begin{document}

\title{Interpreting learning in biological neural networks as zero-order optimization method}

\author{Johannes Schmidt-Hieber\footnote{University of Twente, Drienerlolaan 5, 7522 NB Enschede, The Netherlands. \newline {\small {\em Email:} \texttt{a.j.schmidt-hieber@utwente.nl}} \newline This work has tremendously profited from several discussions with Wouter Koolen. The author is moreover extremely grateful for helpful suggestions and several interesting remarks that were brought up by Matus Telgarsky. The research has been supported by the NWO/STAR grant 613.009.034b and the NWO Vidi grant VI.Vidi.192.021.}}

\date{}
\maketitle

\begin{abstract}
Recently, significant progress has been made regarding the statistical understanding of artificial neural networks (ANNs). ANNs are motivated by the functioning of the brain, but differ in several crucial aspects. In particular, the locality in the updating rule of the connection parameters in biological neural networks (BNNs) makes it biologically implausible that the learning of the brain is based on gradient descent. In this work, we look at the brain as a statistical method for supervised learning. The main contribution is to relate the local updating rule of the connection parameters in BNNs to a zero-order optimization method. It is shown that the expected values of the iterates implement a modification of gradient descent. 

\end{abstract}

%
%

\section{Introduction}

Compared to artificial neural networks (ANNs), the brain learns faster, generalizes better to new situations and consumes much less energy. A child only requires a few examples to learn to discriminate a dog from a cat. And people only need a few hours to learn how to drive a car. AI systems, however, need thousands of training samples for image recognition tasks. And the self-driving car is still under development, despite the availability of data for millions of kilometers of test drives and billions of kilometers of simulated drives. The superhuman performance of AI for some tasks \cite{SCHMIDHUBER201585, SilverEtAl, Brown2018} has to be related to the huge databases and the enormous computing power required for the training.

\begin{figure}
\begin{center}
\begin{center}
	\includegraphics[scale=0.4]{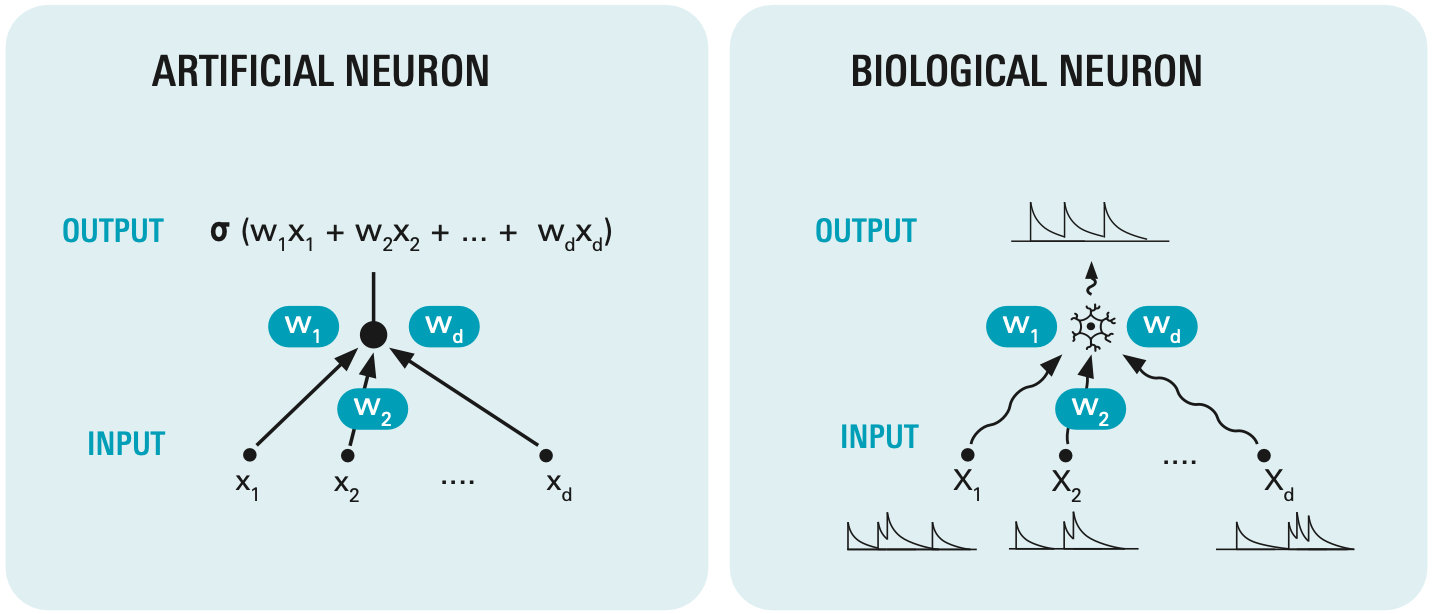} 
\end{center}
\caption{\sl \label{fig.NN}  Artificial neurons receive and output numbers while biological neurons receive and output spike trains.}
\vspace{-0.1cm}
\end{center}
\end{figure}

When identifying the causes for the differences in statistical behavior, it is important to emphasize that although ANNs are inspired by the functioning of the brain, they are very different from biological neural networks (BNNs). Each biological neuron emits a so called spike train that can be modelled as a stochastic process or, more precisely, as a point process \cite{MR1950431, MR2371524} and all computations in BNNs, including the updating of the network parameters, are local. The signal in ANNs, however, is passed instantaneously through the whole network without a time component such as a spike train structure. In conclusion, ANNs generate functions and BNNs point processes.

Another difference between ANNs and BNNs is the learning. Fitting the network parameters in large ANNs is based on variations of stochastic gradient descent (SGD) using the backpropagation algorithm. The parameter update at each network weight is global in the sense that every component of the gradient depends, in general, on all the other, possibly millions of network weights in the whole network. This means that SGD methods require knowledge of the state of the whole network to update one parameter. This is also known as the weight transportation problem \cite{GROSSBERG198723}. As neurons in a biological network do not have the capacity to transport all the information about the state of the other weights, learning in BNNs cannot be driven by gradient descent \cite{Lillicrap2020}. In \cite{Crick1989}, Francis Crick writes: "Nevertheless, as far as the learning process is concerned, it is unlikely that the brain actually uses backpropagation."


In this work, we link the local updating rule for the parameters in a BNN to a derivative-free (or more specifically, a zero-order) optimization method that does not require evaluation of the gradient. Theorem \ref{thm.GD_rewritten} shows that, in expectation, this scheme does approximately gradient descent. 


\section{A brief introduction to biological neural networks (BNNs)} 
\label{sec.BNNs}

\begin{floatingfigure}[r]{6cm}
\begin{framed}
\vspace{-0.4cm}
\begin{center}
\begin{center}
	\includegraphics[scale=0.25]{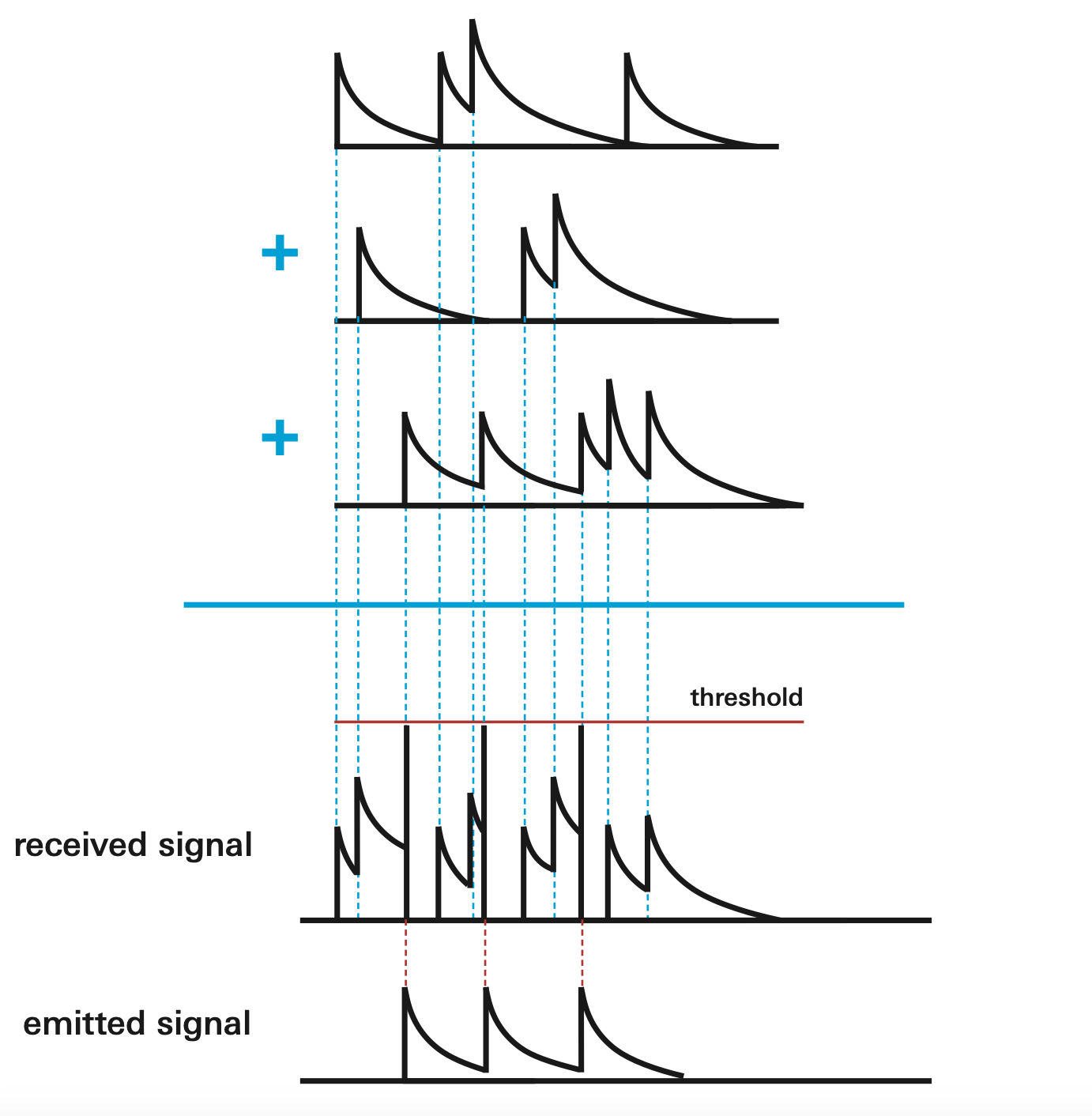} 
\end{center}
\vspace{-0.2cm}
\caption{\sl \label{fig.superp}  Receiving three spike trains, the biological neuron superimposes them and releases spikes, whenever the threshold value (in red) is exceeded.}
\vspace{-0.25cm}
\end{center}
\end{framed}
\end{floatingfigure}

Using graph theory terminology, a BNN is a directed graph, with nodes representing neurons. The nodes/\newline neurons can receive spikes via incoming edges and emit spikes via outgoing edges. In the directed graph, parent nodes are also called presynaptic neurons and children nodes are called postsynaptic neurons. A simple, first model is to think of a spike that is emitted at time $\tau$ as a signal or function $t\mapsto e^{\tau-t}\mathbf{1}(t\geq \tau)$ with $\mathbf{1}(\cdot)$ the indicator function. If neuron $i$ emits a spike at time $\tau$ and is connected to neuron $j$, then neuron $j$ receives the signal $w_{ij}e^{\tau-t}\mathbf{1}(t\geq \tau),$ where $w_{ij}$ is the weight parameter measuring the strength of the connection between the neurons $i$ and $j.$ Due to the exponential decay, the signal fades out quickly. 

When does neuron $j$ fire/emits a spike? Suppose neuron $j$ has incoming edges from neurons $i_1,\ldots,i_m.$ These neurons will occasionally send spikes to $j$ and the overall received signal/potential at $j$ is the superposition of the weighted incoming signals. If the combined signal exceeds a threshold $S$, $j$ fires and all children nodes (postsynaptic neurons) of $j$ receive a signal from $j.$ The generation of the spike trains is illustrated in Figure \ref{fig.superp} for $m=3.$ The three incoming spike trains were chosen for illustrative purposes as triggering a spike in a BNN requires between 20 to 50 incoming spikes within a short time period \cite{gerstner_kistler_naud_paninski_2014}, p.10. After a spike, neuron $j$ enters a short rest phase before it gets back to its normal state. Although this rest phase might play a role in the learning, it will be ignored in the analysis.

The parameters in the BNN are the non-negative weights measuring the strength of the connections. Plasticity is the neuroscience term to describe the changes of the network parameters. Spike time dependent plasticity (STDP) predicts that the parameter $w_{ij}$ measuring the signal strength between the neurons $i$ and $j$ is decreased if a spike is sent from $i$ to $j$ and increased if neuron $j$ emits a spike \cite{Makram97, Bi10464, ZhangNature98, Song2000}. The increase becomes bigger if the time lag between the arrived spike and the firing of neuron $j$ gets smaller. This is known as "fire together, wire together" and is the main principle underlying Hebbian learning \cite{HebbBook, 10.2307/24939213}. 

Among specific forms for the updating formula, as simple but realistic model is to assume that if the spike from neuron $i$ to neuron $j$ arrives at time $\tau$, the weight is decreased by $A_-(w_{ij}) Ce^{- c(\tau-T_-)}$ at time $\tau$ and increased by $A_+(w_{ij}) Ce^{- c(T_+-\tau)}$ at time $T_+,$ where $c, C$ are constants and $T_-,T_+$ are the last/first spike time of neuron $j$ before/after $\tau.$ Regarding the amplitude functions, $A_{\pm}(w_{ij}),$ different choices are possible, see Section 19.2.2 in \cite{gerstner_kistler_naud_paninski_2014} From now on we will study the case that $A_{\pm}(x)=x.$ For $C\leq 1,$ this choice guarantees that the change of the weight is always smaller than the weight itself. Thus positive weights remain positive and the network topology does not change during learning. Combining both updating steps into one formula, we have 
\begin{align}
    w_{ij}  \leftarrow w_{ij}+w_{ij}C(-e^{- c(\tau-T_-)}+e^{- c(T_+-\tau)}).
    \label{eq.update_unsupervised}
\end{align}

As a reward for how well the task has been completed compared to earlier trials and also accounting for the total number of trials, a neurotransmitter such as dopamine is released. The higher the reward, the more the parameters are changed. If the brain performed poorly in the past and suddenly manages to solve a task well, much more neurotransmitter is released than if the same task has already been completed equally well in the past. To take this into account, it has been argued in the neural coding literature that the realized reward is the objective reward, how well the task has been completed, minus the expected reward measuring how well the brain anticipated to do this task \cite{Fremaux13326}. Denote by $R$ the reward and let $\overline R$ be a measure for the anticipated reward. The reward-based synaptic plasticity updating rule becomes then
\begin{align}
	w_{ij}  \leftarrow w_{ij}+(R-\ol R)w_{ij}C\Big(-e^{- c(\tau-T_-)}+e^{- c(T_+-\tau)}\Big).
	\label{eq.updating_HL}
\end{align}
The reward is only released after the prediction has been made. In the meantime, several spikes could have been sent from neuron $i$ to neuron $j.$ This requires that the system has a short-term memory, \cite{Pawlak2010}.

If the brain has to complete a similar task more frequently, it becomes less exciting over time, resulting in a smaller reward. This can be incorporated into the dynamics by including a learning rate $\alpha>0$,
\begin{align}
	w_{ij}  \leftarrow w_{ij}+\alpha (R-\ol R)w_{ij}C\Big(-e^{- c(\tau-T_-)}+e^{- c(T_+-\tau)}\Big).
	\label{eq.updating_HL2}
\end{align}
Supervised learning is more commonly formulated in loss functions than rewards. Because a high reward corresponds to a small loss and vice versa, $L:=-R$ is a loss function, $\ol L=-\ol R$ is the anticipated loss, and the updating formula becomes
\begin{align}
	w_{ij}  \leftarrow w_{ij}+\alpha (L-\ol L) w_{ij}C\Big(e^{- c(\tau-T_-)}-e^{- c(T_+-\tau)}\Big).
	\label{eq.updating_HL_loss}
\end{align}
A key observation is that these updating formulas are derivative-free in the sense that they involve the reward (or loss) but not its gradient.

Hebbian learning rules, such as \eqref{eq.updating_HL_loss}, model the updating of individual weights, but do not explain how the brain can learn a task. A brief overview about relevant existing ideas on learning in BNNs is given in Section \ref{sec.lit}

\bigskip

\section{Zero-order optimization}

Suppose we want to fit a $d$-dimensional parameter vector $\btheta$ to the data and write $L(\btheta)$ for the (training) loss incurred by parameter $\btheta.$ Derivative-free optimization procedures do not require computation of the gradient of the loss. A simple iterative derivative-free scheme would be to randomly pick in each round a new candidate parameter and update the parameter if the loss is decreased. Standard references for derivative-free optimization include \cite{MR1968388, MR2487816, Duchi2015, MR3963507, Primer20}.

Zero-order methods (sometimes also called zero-th order methods) are specific derivative-free optimization procedures. To explain the concept, recall that standard gradient descent is an iterative procedure aiming to minimize the loss function $\btheta\mapsto L(\btheta)$ by the iterative scheme
\begin{align*}
	\btheta_{k+1}=\btheta_k -\alpha_{k+1} \nabla L(\btheta_k), \quad k=0,1,\ldots
\end{align*}
where the initial values $\btheta_0$ are chosen in some way, $\alpha_{k+1}>0$ is the learning rate and $\nabla L(\btheta_k)$ denotes the gradient of the loss function at $\btheta_k.$ In contrast, zero-order methods are only allowed to access the loss function but not the gradient of the loss. From the loss, one can build, however, an estimator for the gradient of the loss. $1$-point zero-order methods replace $-\nabla L(\btheta_k)$ by 
\begin{align*}
	\beta L(\btheta_k+\bxi_k)\bxi_k
\end{align*}
with $\bxi_k$ a $d$-dimensional random vector and $\beta$ a constant. To see how this relates to the gradient, consider the specific case that $\bxi_k$ is multivariate normal with zero mean vector and covariance matrix $\sigma^2I_d,$ where $I_d$ denotes the $d\times d$ identity matrix. The multivariate version of Stein's lemma \cite{MR2724359} states that 
\begin{align}
    \E[L(\btheta_k+\bxi_k)\bxi_k]=\sigma^2\E[\nabla L(\btheta_k+\bxi_k)]  
    \label{eq.Stein}
\end{align}
under weak regularity conditions ensuring that all expectations are well-defined. This means that $\sigma^{-2} L(\btheta_k+\bxi_k)\bxi_k$ estimates the gradient at $\btheta_k+\bxi_k$, that is, $\nabla L(\btheta_k+\bxi_k)=\nabla L(\btheta_k)+$error$_k.$ The hope is that over many iterations the noise contributions cancel out such that in the long-run, the $1$-point zero-order dynamics behaves similarly as gradient descent. The argument above can be extended to general symmetric distributions of $\bxi_k$ that are not necessarily Gaussian. 

Unfortunately, the variance of the $1$-point zero-order gradient estimator \eqref{eq.Stein} can be extremely large and often scales quadratically in the number of parameters $d.$ As an example, suppose that the data are stored in a $d$-dimensional vector $\bY=(Y_1,\ldots,Y_d)^\top$ and consider the least squares loss $L(\btheta)=\|\bY-\btheta\|_2^2.$ Taking $\bxi_k=(\xi_{k1},\ldots,\xi_{kd})\sim \mN(0,\sigma^2I_d)$ and $\beta=\sigma^{-2}$, as above, we have for the $j$-th component of $\beta L(\btheta_k+\bxi_k)\bxi_k$ that 
\begin{align*}
	\sigma^{-2} \big\|\bY-\btheta_k-\bxi_k\big\|_2^2 \xi_{kj}
	= \sigma^{-2} \big(Y_j-\theta_{kj}-\xi_{kj}\big)^2  \xi_{kj}
	+ \sigma^{-2}\sum_{\ell:\ell \neq j} \big(Y_\ell-\theta_{k\ell}-\xi_{k\ell}\big)^2  \xi_{kj}.
\end{align*}
The second term on the right hand side has zero mean. It is pure noise and does not help to estimate the gradient. This sum is over $d-1$ summands and its variance scales with $O(d^2)$ in the number of parameters $d.$

The large noise helps to avoid local minima but can also lead to scenarios for which $1$-point zero-order dynamics quickly diverges to infinity. Indeed if one iterate $\btheta_k$ is already far away from the minimum, the large loss can result in a parameter update $\btheta_{k+1}$ which is much further away from the minimizer than $\btheta_k,$ leading to an even larger loss and an exponential growth of the loss as the number of iterations is further increased.

Regarding theory of zero-order methods, \cite{Duchi2015} studies a related zero-order methods and mirror descent. Assuming that the parameter vector lies in an Euclidean ball, they obtain in their Corollary 1 the rate $\sqrt{d/k}$ with $k$ the number of iterations and also provide a corresponding lower bound proving that this rate is optimal (their Proposition 1). The large noise causes the factor $\sqrt{d}$ in the rate, suggesting slow convergence in the high-dimensional regime. \cite{MR3627456} also finds a suboptimality of order $d$ if zero-order methods are compared to gradient descent. Table 1 in \cite{Primer20} shows that the factor $\sqrt{d}$ or $d$ occurs in all known convergence rates unless second-order information is used.


Due to the large noise, derivative-free methods are in general thought to be inferior compared to gradient descent. This is for instance remarked in \cite{MR2487816}, Section 1.3: "Finally, we want to make a strong statement that often councils against the use of derivative-free methods: if you can obtain clean derivatives (even if it requires considerable effort) and the functions defining your problem are smooth and free of noise you should not use derivative-free methods." 

Zero-order methods are also not necessarily much faster to compute than gradient descent iterates. For the gradient-based backpropagation of ANNs, the number of operations required for the forward pass is of the same order as the number of operations required for the backwards pass. Evaluation of the loss is therefore not substantially cheaper than computing the gradient and zero-order methods cannot be computed at a faster order than backpropagation.

Despite these rather discouraging remarks, there is a rapidly increasing interest in derivative-free methods and they are successfully applied in practice, for example by Google \cite{46180}.

\section{Hebbian learning as zero-order optimization method}

The updating formula \eqref{eq.updating_HL_loss} allows to address supervised learning tasks, where we want to learn the functional relationship between inputs and outputs given observations (or training data) from input-output pairs $(\bX_1,Y_1),(\bX_2,Y_2),\ldots$ that are all generated from the same, unknown distribution as the vector $(\bX,Y).$ Well-known examples for this framework are classification and regression. For instance to classify cat and dog images, $\bX_i$ is the $i$-th image containing all the pixel values of the $i$-th cat image and $Y_i$ is the corresponding label "cat" or "dog", coded as $0$ or $1.$ 

Consider now a feedforward biological neural network (BNN) with $m$ neurons. This means that the neurons/nodes form a directed acyclic graph (DAG) with input neurons receiving information from the data $\bX_i$ and possibly several output neurons. For the subsequent analysis, we neither have to specify a layered structure as commonly done for ANNs nor conversion rules how vector valued inputs are converted into spike trains or output spike trains are cast into response variables, such as conversion into labels in a classification problem. 

In the $k$-th instance, we feed the $k$-th input vector $\bX_k$ in the BNN, let the BNN run and receive then as output the predicted response $\wh Y_k.$ The loss at this round is a measure for the difference between the predicted response $\wh Y_k$ and the real response $Y_k.$ It will be denoted by $L(\wh Y_k,Y_k)$ in the following. The anticipated loss that occurs in \eqref{eq.updating_HL_loss} could be modelled by a (weighted) average over past iterations. Here we use the loss of the previous iterate $L(\wh Y_{k-1},Y_{k-1}).$

During each instance, several spikes can be sent between any two connected neurons. We impose the (strong) assumption that for every run, and any connection, exactly one spike will be released.

Number the $m$ nodes, that represent the neurons in the graph, by $1,\ldots,m$ and denote the edge set by $\mT.$ A pair $(i,j)$ is in $\mT$ if and only if neuron $i$ is a presynaptic neuron for neuron $j.$ Equivalently, $(i,j)\in \mT$ iff there is an arrow from $i$ to $j$ in the underlying DAG. We consider the case that the BNN topology is static, that is, the edge set $\mT$ does not change during learning.

If $w_{ij}^{(k)}$ is the BNN weight after the $k$-th round, it is then updated in the $(k+1)$-st iteration according to \eqref{eq.updating_HL_loss} 
\begin{align}
    &w_{ij}^{(k+1)} \label{eq.wijk_interm1} \\
    &=w_{ij}^{(k)}+\alpha_{k+1}\big( L(\wh Y_k,Y_k)-  L(\wh Y_{k-1},Y_{k-1})\big) w_{ij}^{(k)} C\Big(
    e^{-c(\tau_{ij}^{(k)}-T_{-,j}^{(k)})}
    -e^{-c(T_{+,j}^{(k)}-\tau_{ij}^{(k)})}
    \Big),\notag 
\end{align}
for all $(i,j)\in \mT$ and $\alpha_{k+1}>0$ the learning rate. Here $T_{-,j}^{(k)}$ and $T_{+,j}^{(k)}$ are the closest spike times of the $j$-th neuron before/after the arrival time $\tau_{ij}^{(k)}$ of the spike that is sent from neuron $i$ to neuron $j.$ The constant $C$ can be integrated into the loss function and is from now on set to one. 

For the updating, the location of $\tau_{ij}^{(k)}$ is important within the interval $[T_{-,j}^{(k)},T_{+,j}^{(k)}],$ while the interval length seems to play a minor role. Therefore, we assume that the interval length is constant and set $A:=(T_{+,j}^{(k)}-T_{-,j}^{(k)})/2.$ We assume moreover that the arrival time of the spike from neuron $i$ to neuron $j$ has a negligible influence on the spike times of neuron $j$, that the spike times $\tau_{ij}^{(k)}$ are all independent of each other, and follow a uniform distribution on the interval $[T_{-,j}^{(k)},T_{+,j}^{(k)}].$ As mentioned before, to trigger a spike, it needs of the order of $20-50$ presynaptic neurons to fire in a short time interval. The influence of an individual neuron seems therefore rather minor, justifying the previous assumption. The assumptions above show that the random variable $U_{ij}^{(k)}:=\tau_{ij}^{(k)}-\tfrac 12 (T_{+,j}^{(k)}+T_{-,j}^{(k)})$ are jointly independent and uniformly distributed on $[-A,A].$ Hence, \eqref{eq.wijk_interm1} becomes
\begin{align*}
    w_{ij}^{(k+1)}=w_{ij}^{(k)}+\alpha_{k+1} \big( L(\wh Y_k,Y_k)- L(\wh Y_{k-1},Y_{k-1})\big) w_{ij}^{(k)}\Big(
    e^{-c(A+U_{i,j}^{(k)})}
    -e^{-c(A-U_{i,j}^{(k)})}
    \Big),
\end{align*}
for all $(i,j)\in \mT.$ The factor $e^{-cA}$ can be absorbed into the loss function and the constant $c$ can be absorbed into the hyperparameter $A.$ By reparametrization, we obtain the updating formula
\begin{align}
    w_{ij}^{(k+1)}=w_{ij}^{(k)}+\alpha_{k+1}\big( L(\wh Y_k,Y_k)- L(\wh Y_{k-1},Y_{k-1})\big) w_{ij}^{(k)}\Big(
    e^{-U_{i,j}^{(k)}}
    -e^{U_{i,j}^{(k)}}
    \Big),
    \label{eq.update_recalibrated}
\end{align}
for all $(i,j)\in \mT.$

To further analyze this scheme, it is important to understand how the predicted response $\wh Y_k$ depends on the parameters. We now argue that, under the same assumptions as before, $\wh Y_k$ is a function of the variables $w_{ij}^{(k)}e^{U_{i,j}^{(k)}}.$ The high-level rationale is that in this neural model, all the information that is further transmitted in the BNN about the parameter $w_{ij}^{(k)}$ sits in the spike times of neuron $j$ and the interarrival spike times only depend on $w_{ij}^{(k)}$ through $w_{ij}^{(k)}e^{U_{i,j}^{(k)}}.$ To see this, fix neuron $j.$ The only information that this node/neuron releases to its descendants in the DAG are the spike times of this neuron. This means that from all the incoming information that neuron $j$ receives from presynaptic neurons (parent nodes) only the part is transmitted that affects the spike times of neuron $j.$ As mentioned in Section \ref{sec.BNNs}, a spike arriving at neuron $j$ from neuron $i$ at time $\tau_{ij}^{(k)}$ causes the potential $t\mapsto w_{ij}^{(k)}e^{\tau_{ij}^{(k)}-t}\mathbf{1}(t\geq \tau_{ij}^{(k)})$ at node $j.$ If every incoming neuron spikes once, the overall potential of neuron $j$ is $\sum_{i:(i,j)\in \mT} w_{ij}^{(k)}e^{\tau_{ij}^{(k)}-t}\mathbf{1}(t\geq \tau_{ij}^{(k)}).$ If $S$ denotes the threshold value for the potential at which a neuron spikes, then at the spike time $T_{+,j}^{(k)}$ of the $j$-th neuron, we have by the definition of $U_{ij}^{(k)},$ $S=\sum_{i:(i,j)\in \mT} w_{ij}^{(k)}e^{\tau_{ij}^{(k)}-T_{+,j}^{(k)}}=\sum_{i:(i,j)\in \mT} w_{ij}^{(k)}e^{U_{ij}^{(k)}-\tfrac 12 (T_{+,j}^{(k)}-T_{-,j}^{(k)})}.$ Rearranging this equation shows that the interarrival spike time $T_{+,j}^{(k)}-T_{-,j}^{(k)}$ can be expressed in terms of the variables $w_{ij}^{(k)}e^{U_{ij}^{(k)}}.$ Introduce $\bw_k:=(w_{ij}^{(k)})_{(i,j)\in \mT},$ $\bU_k:=(U_{ij}^{(k)})_{(i,j)\in \mT}$ and write $\bw_k e^{\bU_k}$ for $(w_{ij}^{(k)}e^{U_{i,j}^{(k)}})_{(i,j)\in \mT}.$ The previous argument indicates that the predictor $\wh Y_k$ is a function of $\bw_k e^{\bU_k}$ and $\bX_k.$ Thus, the loss $L(\wh Y_k,Y_k)$ can be written as a function of the form $L\big(\bw_k e^{\bU_k},\bX_k,Y_k\big)$ and \eqref{eq.update_recalibrated} becomes
\begin{align}
    &w_{ij}^{(k+1)} \label{eq.update_recalibrated2} \\
    &=w_{ij}^{(k)}+\alpha_{k+1}\Big( L\big(\bw_k e^{\bU_k},\bX_k,Y_k\big)-L\big(\bw_{k-1} e^{\bU_{k-1}},\bX_{k-1},Y_{k-1}\big)\Big) w_{ij}^{(k)}\Big(
    e^{-U_{i,j}^{(k)}}
    -e^{U_{i,j}^{(k)}}
    \Big). \notag
\end{align}
In a BNN, the parameters $w_{ij}^{(k)}$ are non-negative. We now introduce the real-valued variables $\theta_{ij}^{(k)}=\log(w_{ij}^{(k)})$ and $\btheta_k=(\theta_{ij}^{(k)})_{(i,j)\in \mT}.$ This means that $w_{ij}^{(k)}=e^{\theta_{ij}^{(k)}}.$ A first order Taylor expansion shows that for real numbers $u,v, \Delta$ such that $e^{-v}\Delta$ is small, $e^u=e^v+\Delta$ gives $u=\log(e^v+\Delta)=v+\log(1+e^{-v}\Delta)\approx v+e^{-v}\Delta.$ Working with this approximation, we can rewrite the formula \eqref{eq.update_recalibrated2} in terms of the $\theta$'s as 
\begin{align}
    &\theta_{ij}^{(k+1)} \label{eq.update_theta_1} \\
    &=\theta_{ij}^{(k)}+\alpha_{k+1}\Big( L\big(\btheta_k+\bU_k,\bX_k,Y_k\big)-L\big(\btheta_{k-1}+\bU_{k-1},\bX_{k-1},Y_{k-1}\big)\Big)\Big(
    e^{-U_{i,j}^{(k)}}
    -e^{U_{i,j}^{(k)}}
    \Big).\notag
\end{align}
Relating this formula to gradient descent and the weight transportation problem mentioned in the introduction, we see that the update of one parameter only depends on all the other parameters through the value of the loss function.

In vector notation, the previous equality becomes
\begin{align}
    &\btheta_{k+1} \label{eq.updating}	\\
    &=\btheta_k+\alpha_{k+1} \big( L(\btheta_k+ \bU_k,\bX_k,Y_k)-L(\btheta_{k-1}+ \bU_{k-1}, \bX_{k-1},Y_{k-1}) \big)\big(e^{-\bU_k}-e^{\bU_k}\big),
	\notag
\end{align}
where  $e^{\bU_k}$ and $e^{-\bU_k}$ should be understood as componentwise applying the functions $x\mapsto e^x$ and $x\mapsto e^{-x}$ to the vector $\bU_k.$ In particular, the loss is always a scalar and $e^{\bU_k},$ $e^{-\bU_k}$ are $d$-dimensional vectors.

So far, we have not specified any initial conditions. From now on, we assume that the initial values $\btheta_0, \btheta_{-1}$ are given and that all the other parameter updates are determined by \eqref{eq.updating} for $k=0,1,\ldots$ with $\bU_{-1}, \bU_0,\bU_1,\bU_2,\ldots$ drawn i.i.d. from the uniform distribution $\mU([-A,A]^d).$

For the derivation of \eqref{eq.updating}, we chose the loss of the previous iterate $L(\wh Y_{k-1},Y_{k-1})$ as anticipated loss. This results in the term $L(\btheta_{k-1}+ \bU_{k-1}, \bX_{k-1},Y_{k-1})$ in \eqref{eq.updating}. If we would instead consider a weighted average $\sum_{\ell\geq 1} \gamma_\ell L(\wh Y_{k-\ell},Y_{k-\ell})$, we get as update equation
\begin{align*}
   \btheta_{k+1}
    =\btheta_k+\alpha_{k+1} \Big( L(\btheta_k+ \bU_k,\bX_k,Y_k)-\sum_{\ell\geq 1} \gamma_\ell L(\btheta_{k-\ell}+ \bU_{k-\ell}, \bX_{k-\ell},Y_{k-\ell}) \Big)\big(e^{-\bU_k}-e^{\bU_k}\big).
\end{align*}
It is natural to assume that the weights should sum up to one and that less weight is given to instances that happened further in the past, that is, $\gamma_1\geq \gamma_2\geq \ldots.$ Possible forms of discounting for the past include exponential weight decrease $\gamma_\ell \propto e^{-\lambda \ell}$ and polynomial weight decrease $\gamma_\ell \propto \ell^{-\lambda},$ with $\lambda$ a positive parameter and proportionality constants determined by the normalization condition $\sum_{\ell \geq 1} \gamma_\ell =1.$

As an analogue of \eqref{eq.Stein}, the next result shows that in average, this dynamic can also be understood as a gradient descent method with gradient evaluated not exactly at $\btheta_k$ but at a random perturbation $\btheta_k+\bU_k.$

\begin{thm}
\label{thm.GD_rewritten}
Write $\bU_k=(U_{k1},\ldots,U_{kd})^\top$ and let $e^A-e^{\bU_k}$ and $e^A-e^{-\bU_k}$ be the $d$-dimensional vectors with components $e^A-e^{U_{kj}}$ and $e^A-e^{-U_{kj}},$ respectively. Denote by $\odot$ the Hadamard product (componentwise product) of two matrices/vectors of the same dimension(s). Assuming that the gradient of $L$ exists, we have
\begin{align}
	\E\big[\btheta_{k+1}\big]=
	\E\big[\btheta_k\big]
	-
	\alpha_{k+1} e^{-A} \E\Big[ \nabla_{\btheta_k} L(\btheta_k+ \bU_k,\bX_k,Y_k) \odot 
	\big(e^A-e^{\bU_k}\big) \odot \big(e^A-e^{-\bU_k}\big)\Big].
	\label{eq.gradient_descent_rewritten}
\end{align}
\end{thm}

Instead of taking the expectation over all randomness, the statement is also true if we only take the expectation with respect to $\bU_k,$ which is the same as the conditional expectation $\E[\cdot |\bU_{-1},\bU_0,\bU_1,\ldots,\bU_{k-1}, (\bX_\ell,Y_\ell)_{\ell \geq 1}].$

Note that $(e^A-e^{U_{kj}})(e^A-e^{-U_{kj}})$ is non-negative. Thus $f_A(x) =C(A)^{-1}(e^A-e^x)(e^A-e^{-x})\mathbf{1}(-A\leq x\leq A)$ defines a probability density function for the positive normalization constant $C(A)=2A(e^{2A}+1)+2-2e^{2A}=\int_{-A}^A (e^A-e^x)(e^A-e^{-x})\, dx$. Denoting by $\partial_j L(\bv,\bX_k,Y_k)$ the partial derivative of $L$ with respect to the $j$-th component of $\bv,$ we can state the previous result componentwise as
\begin{align}
	\E\big[\btheta_{k+1,j}\big]=
	\E\big[\btheta_{kj}\big]
	-
	\alpha_{k+1} e^{-A}C(A) \E\Big[ \partial_j L(\btheta_k+ \bU_k^{(j)},\bX_k,Y_k)\Big],
	\label{eq.gradient_descent_rewritten2}
\end{align}
for a random vector $\bU_k^{(j)}=(U_{k1},\ldots,U_{k,j-1},V_{kj},U_{k,j+1},\ldots,U_{kd})^\top,$ with jointly independent random variables $V_{kj}\sim f_A$ and $U_{k\ell} \sim \mU[-A,A],$ $\ell=1,\ldots,j-1,j+1,\ldots,d.$

\begin{proof}[Proof of Theorem \ref{thm.GD_rewritten}]
Similarly as Stein's identity \eqref{eq.Stein}, Equation \eqref{eq.gradient_descent_rewritten} can be proved using integration by parts. Throughout the proof, we omit the dependence of the loss function $L$ on the data. By conditioning on $(\bU_{-1},\bU_0,\ldots,\bU_{k-1}, (\bX_\ell,Y_\ell)_{\ell \geq 1})$ and the fact that $e^{-\bU_k}$ and $e^{\bU_k}$ have the same distribution, it follows that 
\begin{align}
\begin{split}
	&\E\Big[ L(\btheta_{k-1}+ \bU_{k-1}) \big(e^{-\bU_k}-e^{\bU_k}\big)\Big] \\
	&= 
	\E\Big[ L(\btheta_{k-1}+ \bU_{k-1}) \E\Big[\big(e^{-\bU_k}-e^{\bU_k}\big) \, \Big| \, \bU_{-1},\bU_0,\ldots,\bU_{k-1}, (\bX_\ell,Y_\ell)_{\ell \geq 1}\Big]\Big] \\
	&=0.
\end{split}
\label{eq.2617w}	
\end{align}
With $\bu=(u_1,\ldots,u_d)^\top,$ the $j$-th component of $e^{-A} \E[ \nabla_{\btheta_k} L(\btheta_k+ \bU_k) \odot (e^A-e^{\bU_k})(e^A-e^{-\bU_k})]$ is 
\begin{align*}
	 &\frac{e^{-A}}{(2A)^d} \int_{[-A,A]^d} \partial_j L(\btheta_k+ \bu) 
	\big(e^A-e^{u_j}\big)\big(e^A-e^{-u_j}\big) \, d\bu\\
	&=
	\frac{e^{-A}}{(2A)^d} \int_{[-A,A]^{d-1}}\int_{-A}^A \partial_j L(\btheta_k+ \bu) 
	\big(e^A-e^{u_j}\big)\big(e^A-e^{-u_j}\big) \, du_j du_1 \ldots du_{j-1}d u_{j+1}\ldots du_d,
\end{align*}
Observe that $(e^A-e^{u_j})(e^A-e^{-u_j})$ vanishes at the boundaries $u_j\in \{-A,A\}$ and $\partial_{u_j}(e^A-e^{u_j})(e^A-e^{-u_j})=e^{A-u_j}-e^{A+u_j}.$ Thus, applying integration by parts formula to the inner integral yields 
\begin{align*}
	&\int_{-A}^A \partial_j L(\btheta_k+ \bu) 
	\big(e^A-e^{u_j}\big)\big(e^A-e^{-u_j}\big) \, du_j
	= - 
	e^A \int_{-A}^A L(\btheta_k+ \bu) 
	\big(e^{-u_j}-e^{u_j}\big) \, du_j
\end{align*}
and therefore
\begin{align*}
	 &\frac{e^{-A}}{(2A)^d} \int_{[-A,A]^d} \partial_j L(\btheta_k+ \bu) 
	\big(e^A-e^{u_j}\big)\big(e^A-e^{-u_j}\big) \, d\bu \\
	&= 
	- \frac{1}{(2A)^d} \int_{[-A,A]^d} L(\btheta_k+ \bu) 
	\big(e^{-u_j}-e^{u_j}\big) \, d\bu \\
	&= - \E\Big[L(\btheta_k+ \bU_k)\big(e^{-U_{kj}}-e^{U_{kj}}\big)\Big].
\end{align*}
This holds for all $j=1,\ldots,d.$ The minus on the right hand side cancels out the first minus in \eqref{eq.gradient_descent_rewritten}. Together with \eqref{eq.2617w}, the claim follows.	
\end{proof}

Equation \eqref{eq.2617w} in the proof shows that the theorem still holds if the term $L(\btheta_{k-1}+ \bU_{k-1},\bX_{k-1},Y_{k-1})$ in \eqref{eq.updating} is replaced by zero or any other value that is independent of $\bU_k.$

To obtain a proper zero-order method, a crucial assumption is to choose the amplitude functions $A_+, A_-$ in \eqref{eq.update_unsupervised} to be the same. In the brain, these functions are close, but some authors argue that there is a slight difference \cite{Song2000}. Such differences would lead to additional, small contributions in the iterations that cannot be linked to the gradient.

A statistical analysis of the zero-order method \eqref{eq.updating} is challenging, even for simple models such as data generated from the linear regression model. Another open problem is to determine whether the convergence rate of \eqref{eq.updating} scales in the number of parameters $d$ in the same way as other zero-order methods.

To summarize, we have rewritten the local spike-time dependent plasticity rule for individual weight parameters \eqref{eq.updating_HL_loss} into a zero-order optimization problem \eqref{eq.updating}. While local updating rules cannot implement gradient descent schemes due to the weight transportation problem, Theorem \ref{thm.GD_rewritten} shows that, in average, this optimization scheme implements a modification of gradient descent. This offers a fresh perspective on supervised learning in the brain. It remains to reconcile the observed efficiency of learning in biological neural networks with the slow convergence of zero-order methods.

\section{Literature on learning with BNNs}
\label{sec.lit}

This literature survey is aimed to give a quick overview. For a more detailed summary of related literature, see \cite{TAVANAEI201947, Whittington19}.

To train BNNs on data, a natural idea is to ignore Hebbian learning and to fit BNNs via gradient descent. Similar as backpropagation efficiently computes the gradient in ANNs, SpikeProp \cite{BOHTE200217, BOOIJ2005552} is an algorithm to compute the gradient for spiking neural networks. 

The weight transportation problem is caused by the parameter dependence in the backwards pass of the backpropagation algorithm. Feedback alignment \cite{Lillicrap2016, Nokland2016, Liao_Leibo_Poggio_2016, NEURIPS2018_63c3ddcc, Lillicrap2020} avoids this by using the backpropagation algorithm with random weights in the backwards pass. In a network, the feedback could be then transmitted via specific feedback neurons. 

If the brain does a version of backpropagation, the difficulty is always the feedback from the output backwards to the neurons. Contrastive Hebbian learning \cite{6796552} assumes that there are two different phases. During the first phase the network does prediction and the second phase starts after the prediction error is revealed. In one of the phases the learning is Hebbian and in the other one, the learning is anti-Hebbian. Anti-Hebbian learning means that if two neurons fire together, the connecting weight parameter is decreased instead of increased. Equilibrium propagation \cite{Scellier17} overcomes the two types of learning in the different phases but requires again the computation of a gradient. 

For a biologically more plausible implementation of the weight transportation problem, predictive coding \cite{Whittington17, Whittington19, SongNEURIPS2020, Millidge, Millidge2022} uses two types of neurons, named error nodes and value nodes. These two nodes are associated to each other and process forward and backward information locally.  


\cite{Seung} proposes the concept of a "hedonistic synapse" that follows a Hebbian learning rule and takes the global reward into account. For the learning, a hedonistic synapse has to be able to store information from previous trials in a so-called eligibility trace.

Closest to our approach is weight perturbation \cite{WerfelXS03}. Weight perturbation adds random noise to the parameters or the outputs and compares the loss with and without added noise to estimate the gradient. Whereas the cause of the noise perturbation is not entirely clear in the weight perturbation framework, we have shown in this work, how the spike train structure in BNNs implies a random perturbation of the parameters in the loss with uniformly distributed noise and how this leads to a specific derivative-free updating formula for the weights that also involves the difference of the loss function evaluated for different instance of the noisy parameters.

A more statistical approach is \cite{NIPS2009_a5cdd4aa} considering unsupervised classification using a small BNN. This work identifies a closer link between a Hebbian learning rule and the EM-algorithm for mixtures of multinomial distributions. Some other ideas on unsupervised learning in BNNs are moreover provided in \cite{gerstner_kistler_naud_paninski_2014}, Section 19.3.

To summarize, there are various theories that are centered around the idea that the learning in BNNs should be linked to gradient descent. All of these approaches, however, contain still biological implausibilities and lack a theoretical analysis.

\bibliographystyle{acm}       
\bibliography{bib}           

\end{document}